# Benchmarking AutoML Frameworks for Disease Prediction Using Medical Claims

Roland Albert Romero[1], Mariefel Nicole Y. Deypalan[1], Suchit Mehrotra[1], John Titus Jungao[1], Natalie E. Sheils[1], Elisabetta Manduchi[2] and Jason H. Moore[2*]

[1] OptumLabs at UnitedHealth Group, Minnetonka, MN 55343, USA

[2] Department of Biostatistics, Epidemiology & Informatics and Institute for Biomedical Informatics, Perelman School of Medicine, University of Pennsylvania, Philadelphia, PA 19104, USA

[*] Correspondence to:
Jason H. Moore, PhD, FACMI
Address: D202 Richards Building
3700 Hamilton Walk
University of Pennsylvania
Philadelphia, PA 19104-6116
Email address: jhmoore@upenn.edu
Phone: 215-573-4411

Word Count: 3,199

# ABSTRACT

**Objectives**

Ascertain and compare the performances of AutoML tools on large, highly imbalanced healthcare datasets.

**Materials and Methods**

We generated a large dataset using historical administrative claims including demographic information and flags for disease codes in four different time windows prior to 2019. We then trained three AutoML tools on this dataset to predict six different disease outcomes in 2019 and evaluated model performances on several metrics.

**Results**

The AutoML tools showed improvement from the baseline random forest model but did not differ significantly from each other. All models recorded low area under the precision-recall curve and failed to predict true positives while keeping the true negative rate high. Model performance was not directly related to prevalence. We provide a specific use-case to illustrate how to select a threshold that gives the best balance between true and false positive rates, as this is an important consideration in medical applications.

**Discussion**

Healthcare datasets present several challenges for AutoML tools, including large sample size, high imbalance, and limitations in the available features types. Improvements in scalability, combinations of imbalance-learning resampling and ensemble approaches, and curated feature selection are possible next steps to achieve better performance.

**Conclusion**

Among the three explored, no AutoML tool consistently outperforms the rest in terms of predictive performance. The performances of the models in this study suggest that there may be room for improvement in handling medical claims data. Finally, selection of the optimal prediction threshold should be guided by the specific practical application.

# BACKGROUND AND SIGNIFICANCE

Leveraging big data growth in biomedicine and healthcare, machine learning (ML) has helped improve health outcomes, cut healthcare costs, and advance clinical research [1-4]. Studies applying ML to healthcare data range from models for disease prediction or for improving quality of care, to applications such as detection of claim fraud [2, 5-8]. Clinical big data used in various studies range from electronic health records, medical records, and claims data. Many studies are limited to a single healthcare or hospital system [9-12].

Despite the demonstrated benefits of machine learning, different models need to be trained in the context of the problem to achieve good performance [13]. For each model, domain experts such as clinicians need to collaborate with data scientists to design ML pipelines [14]. Automated machine learning (AutoML) is an emerging field [15] that aims to simplify this labor-intensive process [16] which can accelerate the integration of ML in healthcare scenarios [1]. State-of-the-art AutoML platforms allow domain experts to design decently performing ML pipelines without deep knowledge of ML or statistics while at the same time easing the burden of tedious manual tasks such as model selection and hyperparameter optimization for data scientists [14].

With ML being adopted across industries, standardized benchmarks and datasets are needed to compare competing systems [18]. These benchmark suites need to have datasets that highlight strengths and weaknesses of established ML methods [19]. Despite the emergence of numerous AutoML tools, there is still a need for standardized benchmarks in the field. Multiple studies to benchmark various AutoML tools [14, 20-22] have been done. Notably, [23] presented an open-source AutoML benchmark framework to provide objective feedback on the performance of different AutoML tools. Gijsbers et. al. compared four AutoML tools across 39 public data sets, twenty-two of which are binary classifications, with a mixture of balanced and imbalanced data. Of these, only two have very low prevalence for one class, at around 1.8% each. Most of these studies on benchmarks tested public datasets that have sample sizes in the order $10^3$ and feature sizes between 10-100. In contrast, our study uses a population of over 12 million and over 3,500 features.

A review of published papers for AutoML showed that despite the potential applications and demonstrated need [17], little work has been done in applying AutoML to the field of healthcare [7]. Waring et. al. determined the primary reasons for the lack of AutoML solutions for healthcare to be: (1) the lack of high-quality, representative, and diverse datasets, and (2) the inefficiency of current AutoML approaches for large datasets common in the biomedical environment. In particular, disease prediction problems often involve highly imbalanced datasets [24] which do not lend themselves well to predictive modelling. Disease prevalences are much lower than those of the public datasets used by Gjisbers et.

al; the datasets we consider in this paper have positive prevalence ranging from 0.053% - 0.63%. The extremely low prevalence does not give the models enough samples from one class to train on.

# OBJECTIVE

To advance the use of AutoML tools in healthcare, there is a need to first assess their performance in representative datasets. Doing so brings to light the challenges and limitations of using these tools on healthcare data and serves as the basis for future improvements to better address problems in healthcare. In this study, we generated a dataset using claims data with 12.4M rows and 3.5k features. Using this, we compared the performance of different AutoML tools for predicting outcomes for different diseases of interest on datasets with high class imbalance.

# MATERIALS AND METHODS

## Population

The population used in this analysis consisted of 12,425,832 people insured by a single large national insurer from the UnitedHealth Group Clinical Discovery Portal who were continuously enrolled in Medicare or Commercial plans from January 1, 2018 to December 31, 2019. Continuous enrollment in this period was required since the identification of the disease cohorts and the creation of features are heavily reliant on historic claims data. The prediction period was set as 2019; hence, features were only created using data prior to 2019. While it would have been ideal to ensure the completeness of each person's claims history, imposing a longer continuous enrollment criterion would have made fewer people eligible. Although features were created based on claims data from 2016 to 2018, completeness can only be guaranteed for data in 2018.

## Target Diseases

We aimed to predict the occurrence of six diseases – lung cancer, prostate cancer, rheumatoid arthritis (RA), type 2 diabetes (T2D), inflammatory bowel disease (IBD), and chronic kidney disease (CKD) – in the prediction year. Claims-based definitions were created for each target disease. Table 1 gives definitions for each disease, along with the corresponding prevalence and cohort size, presented in order of increasing prevalence. Disease flags are based on the International Classification of Diseases, Tenth Revision (ICD-10). Since the presence of a given ICD-10 code in a claim may simply be due to an event such as a screening test being ordered rather than truly indicative of a diagnosis, we required the presence of that disease code in at least two claims within a specified time period for most of the diseases under consideration. The second occurrence of the ICD-10 code is considered the confirmatory diagnosis for most diseases.

Table 1. Definitions for flagging disease outcomes and the respective prevalences in the final cohort table. Abbreviations used: Chronic Kidney Disease (CKD), Type 2 Diabetes (T2D), Inflammatory Bowel Disease (IBD), Rheumatoid Arthritis (RA), International Classification of Diseases, Tenth Revision (ICD-10).

| *Disease* | ICD-10 Code | Definition | Prevalence | Number of cases |
|---|---|---|---|---|
| *Lung Cancer* | C34 | Two lung cancer claims at least 30 days apart, no history of any cancer | 0.053% | 6,539 |
| *Rheumatoid Arthritis (RA)* | M05, M06 (Except M064) | At least one RA claim* | 0.10% | 12,174 |
| *Prostate Cancer* | C61 | Two prostate cancer claims at least 30 days apart, no history of any cancer | 0.12% | 14,925 |
| *Type 2 Diabetes (T2D)* | E11 | Two T2D claims at least 30 days apart | 0.59% | 73,540 |
| *Inflammatory Bowel Disease (IBD)* | K51, K52 | Two IBD claims at least one day apart | 0.32% | 39,502 |
| *Chronic Kidney Disease (CKD)* | N18 | Two CKD claims at least 30 days apart | 0.63% | 78,786 |

* RA does not require a confirmatory diagnosis since typically, it only takes one physician visit to diagnose the condition and ICD-10 codes for RA are not typically used to indicate screening exams.

## Data Creation

Features were derived from the claims history of members from 2016 to 2018. Each data record corresponds to one claim, which may be associated with up to 12 different diagnoses. In claims data, the first diagnosis is the principal diagnosis associated with the health service availed, and all other diagnoses that follow are considered secondary. Only the first three diagnoses in each claim were considered in our analysis.

Each diagnosis corresponds to an ICD-10 code, which can be up to 7 digits long. For each of the first three diagnoses, only the first three characters of the ICD-10 codes were used. The presence of these ICD-10 codes in four periods of varying lengths were then flagged. Table 2 shows the time windows considered. Thus, for each of the ICD-10 codes, there are 4 distinct features in our dataset.

Table 2. Time periods for creating feature flags.

| Time window | Start date | End date |
|---|---|---|
| 1 | Oct 1 2018 | Dec 31 2018 |
| 2 | July 1 2018 | Sep 30 2018 |

| 3 | Jan 1 2018 | Jun 30 2018 |
| 4 | Jan 1 2016 | Dec 31 2017 |

Demographic information such as gender, state-level socioeconomic index, and age in 2018 were also used as features in the analysis. In total, 3,511 features were created.

## Benchmark Framework

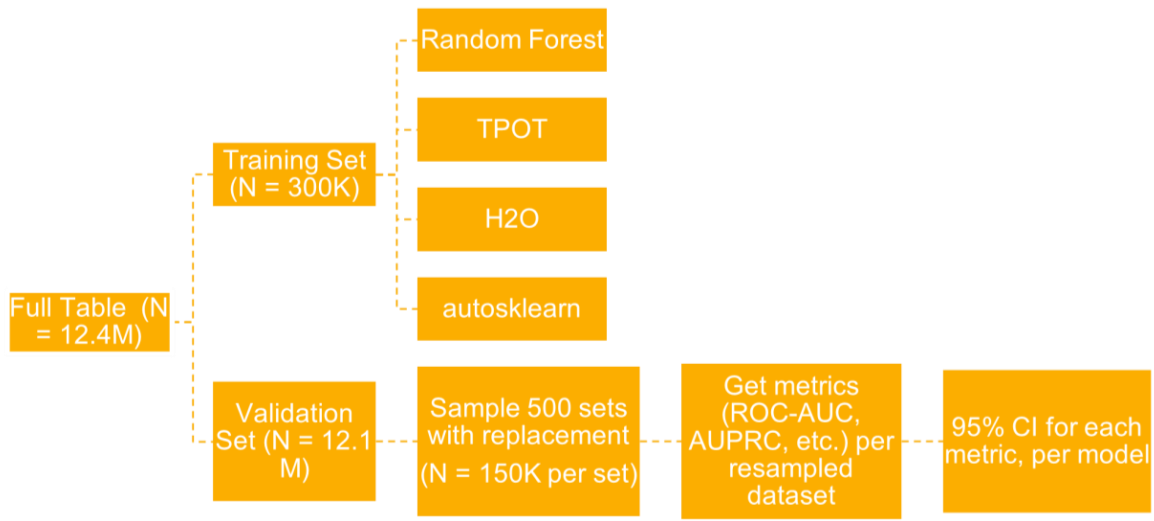

Figure 1. Flowchart for showing framework for benchmarking AutoML tools adapted from Gjisbers et. al.

The flowchart in Figure 1 shows the framework used to benchmark the different AutoML systems adapted from [22] and modified to include a bootstrapping procedure to obtain 95% confidence intervals for each of the metrics considered. The features used for each model depended on the target outcome; flags corresponding to the ICD-10 code of the disease being predicted were excluded. For example, for lung cancer, all four features across different windows for the ICD code C34 were dropped.

For each target disease, we generated a training set of 300,000 samples taken from the population of 12 million, maintaining the disease prevalence. The three AutoML tools (AutoSklearn [25], H2O [26] and TPOT [27]) and a random forest model were trained on the same training set for each disease. Another difference between our framework and [22] is that for each AutoML model, we optimized for different metrics – average precision (area under precision-recall curve (AUCPR) approximation), balanced accuracy, and area under the receiver operating characteristic curve (ROC AUC). H2O was optimized for AUCPR and AUC, the latter corresponding to ROC AUC. We did not optimize H2O for balanced accuracy because this metric was not included in its base built-in scorers. This resulted in multiple models per target disease per tool instead of having a single model optimized for ROC AUC. The random forest model was considered the baseline for comparison. The default settings were used for each tool, except for the maximum run-time which we set at 48 hours for each model. All models were trained on identical 16-CPU 8-core Intel Xeon (2.3 GHz) machines with 256GB RAM. The trained models were then used to predict outcomes for the remaining holdout dataset consisting of 11.7 million samples.

For each model and target disease, bootstrapping was performed on the predictions to obtain 95% confidence intervals for each model metric. Samples were taken with replacement (both stratified and not stratified) from the holdout validation set to obtain 500 sets of 150,000 observations each. Metrics were then computed for the predictions of each model on each resampled dataset, yielding 500 values per metric per model which were used to derive the 95% confidence intervals. We note that, due to the large dataset size and consequent time and resource requirements, we ran each AutoML tool once for each choice of optimization metric, so these are confidence intervals for the performance on the holdout data for each these specific AutoML runs.

## RESULTS

The bootstrapped metrics for the performance of the different models on the holdout set are shown in Figures 2 and 3 for ROC AUC and AUCPR (the latter as approximated by the average precision), respectively. The same results can be seen in tabular form in Tables A1 and A2 in the Appendix.

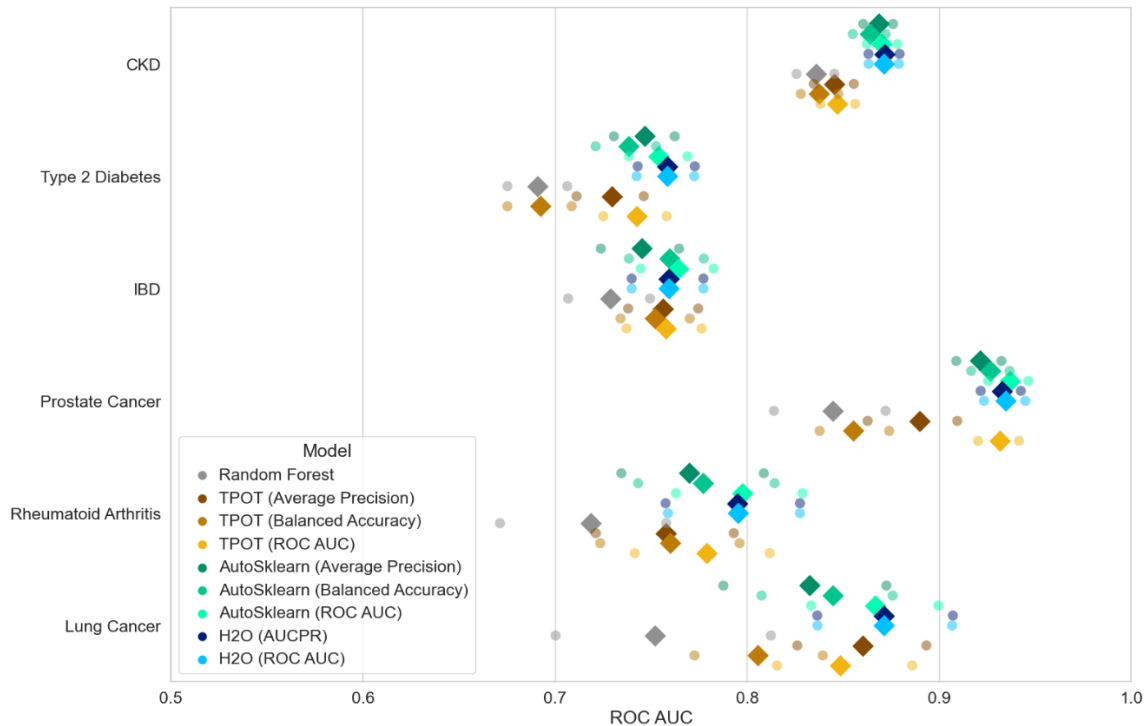

Figure 2. ROC AUC performance of different AutoML models trained for various disease outcomes from stratified bootstrap samples. Median values are indicated by diamond markers and 95% CI limits are indicated by circles.

In both figures, diamond markers indicate the median metric scores for each model, while circle markers denote the lower and upper limits of the 95% confidence intervals calculated through bootstrapping. These figures show metrics computed using stratified bootstrap samples. There is minimal difference between the results of getting the metrics from either stratified or non-stratified bootstrap samples. The results for non-stratified samples can be seen in Figures A1 and A2 in the Appendix.

For ROC AUC, we observe varying performances across different diseases. In general, no single AutoML framework outperforms the rest consistently and with a wide margin. Also, we observe that disease prevalence is not directly correlated to model performance; models with highest ROC AUC scores were those for prostate cancer which is the second least prevalent disease (0.12% prevalence). We also observe narrow confidence intervals for the models trained for predicting CKD, which has the highest prevalence. Wider confidence intervals correspond to lower disease prevalence, with the widest intervals observed for lung cancer (0.053% prevalence). Note that this is not always the case; for prostate cancer, all AutoSklearn and H2O models, and the TPOT model optimized for ROC AUC trained have relatively narrow confidence intervals.

Table 3. Median performance ROC AUC scores for different AutoML models scaled according to median random forest performance. Models with the best performance for each disease are indicated in bold.

| Metric: ROC AUC | Lung Cancer | Prostate Cancer | Rheumatoid Arthritis | Type 2 Diabetes | IBD | CKD |
|---|---|---|---|---|---|---|
| Model | | | | | | |
| AutoSklearn (Average Precision) | 1.107 | 1.091 | 1.072 | 1.081 | 1.022 | 1.039 |
| AutoSklearn (Balanced Accuracy) | 1.124 | 1.097 | 1.082 | 1.069 | 1.042 | 1.034 |
| AutoSklearn (ROC AUC) | 1.152 | **1.109** | **1.110** | 1.091 | **1.048** | 1.041 |
| H2O (AUC) | **1.159** | 1.107 | 1.107 | **1.098** | 1.042 | 1.042 |
| H2O (AUCPR) | **1.159** | 1.104 | 1.106 | **1.098** | 1.042 | **1.043** |
| Random Forest | 1.000 | 1.000 | 1.000 | 1.000 | 1.000 | 1.000 |
| TPOT (Average Precision) | 1.144 | 1.053 | 1.055 | 1.056 | 1.037 | 1.012 |
| TPOT (Balanced Accuracy) | 1.071 | 1.013 | 1.058 | 1.003 | 1.032 | 1.002 |
| TPOT (ROC AUC) | 1.128 | 1.103 | 1.084 | 1.075 | 1.040 | 1.013 |
| Random Forest | 1.000 | 1.000 | 1.000 | 1.000 | 1.000 | 1.000 |

Since model scores and performance varied across diseases, we normalize the median ROC AUC scores based on the median random forest model performance as done by Gjisbers et. al. The results are shown in Table 3. The best performing models across diseases are either H2O models or the

AutoSklearn model optimized for ROC AUC. However, for each disease the difference between the best model the other models are small.

In terms of ROC AUC improvements relative to the random forest models, greater improvements are observed for the less prevalent diseases. The median improvements for all AutoML models per disease are 1.136, 1.100, 1.083, 1.041, 1.078, and 1.036 for lung cancer, prostate cancer, rheumatoid arthritis, IBD, Type 2 Diabetes, and CKD, respectively.

Due to the imbalance of the datasets, we also measure model performance on AUCPR. Low AUCPR scores are observed for all models as seen in Figure 3. The models for prostate cancer which had narrow confidence intervals in terms of their ROC AUC scores have wider confidence intervals for their bootstrapped AUCPR scores. Generally, H2O models had the highest median AUCPR scores. Taking note of the range of AUCPR values, however, there is no single model that outperforms the rest significantly across different diseases.

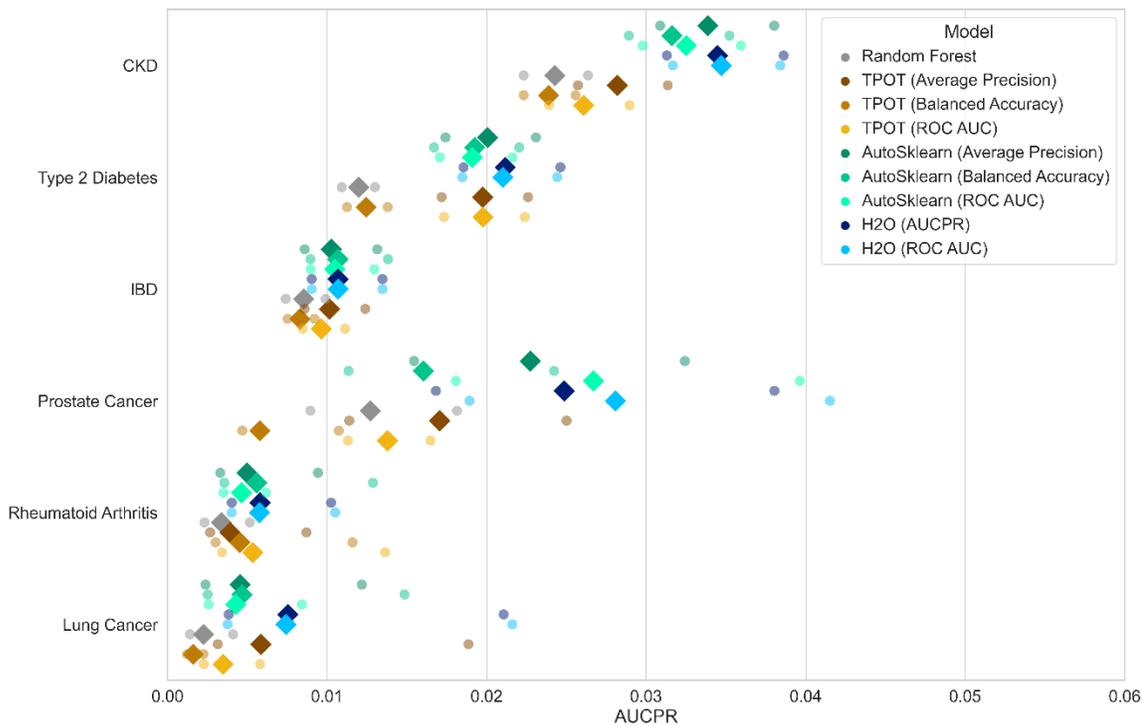

Figure 3. AUCPR performance of different AutoML models trained for various disease outcomes from stratified bootstrap samples. Median values are indicated by diamond markers and 95% CI limits are indicated by circles.

Table 4. Median AUCPR scores for different AutoML models scaled according to median random forest performance. Models with the best performance for each disease are indicated in bold.

| Metric: Average Precision | Lung Cancer | Prostate Cancer | Rheumatoid Arthritis | Type 2 Diabetes | IBD | CKD |
|---|---|---|---|---|---|---|
| Model | | | | | | |
| AutoSklearn (Average Precision) | 1.957 | 1.787 | 1.471 | 1.675 | 1.212 | 1.395 |
| AutoSklearn (Balanced Accuracy) | 2.043 | 1.260 | 1.647 | 1.608 | 1.259 | 1.300 |
| AutoSklearn (ROC AUC) | 1.870 | 2.102 | 1.353 | 1.592 | 1.235 | 1.337 |
| H2O (AUC) | 3.217 | **2.213** | **1.706** | 1.750 | **1.259** | **1.428** |
| H2O (AUCPR) | **3.261** | 1.961 | **1.706** | **1.767** | **1.259** | 1.420 |
| TPOT (Average Precision) | 2.565 | 1.346 | 1.147 | 1.650 | 1.200 | 1.160 |
| TPOT (Balanced Accuracy) | 0.696 | 0.457 | 1.324 | 1.033 | 0.976 | 0.984 |
| TPOT (ROC AUC) | 1.522 | 1.087 | 1.559 | 1.650 | 1.129 | 1.074 |
| Random Forest | 1.000 | 1.000 | 1.000 | 1.000 | 1.000 | 1.000 |

Table 4 shows the performance increases of the models relative to the median baseline scores of the random forest model. Despite the low AUCPR scores, we generally observe improvements in AUCPR compared to the baseline models except for TPOT models optimized for balanced accuracy, especially those trained for predicting prostate and lung cancer. The median AUCPR improvements for all AutoML models per disease are 2.000, 1.567, 1.515, 1.224, 1.650, and 1.319 for lung cancer, prostate cancer, rheumatoid arthritis, IBD, Type 2 Diabetes, and CKD, respectively.

Beyond ROC AUC values, selecting the thresholds for each model is an essential step in evaluating a model for practical purposes. This is especially true when working with imbalanced data [28]. Despite the AutoML output models being ready to generate hard predictions, in practice, one must still consider the threshold that will give the best balance between true positive rate and false positive rate depending on the problem being solved. The actual ROC curves generated using the full validation set are shown in Figure 4.

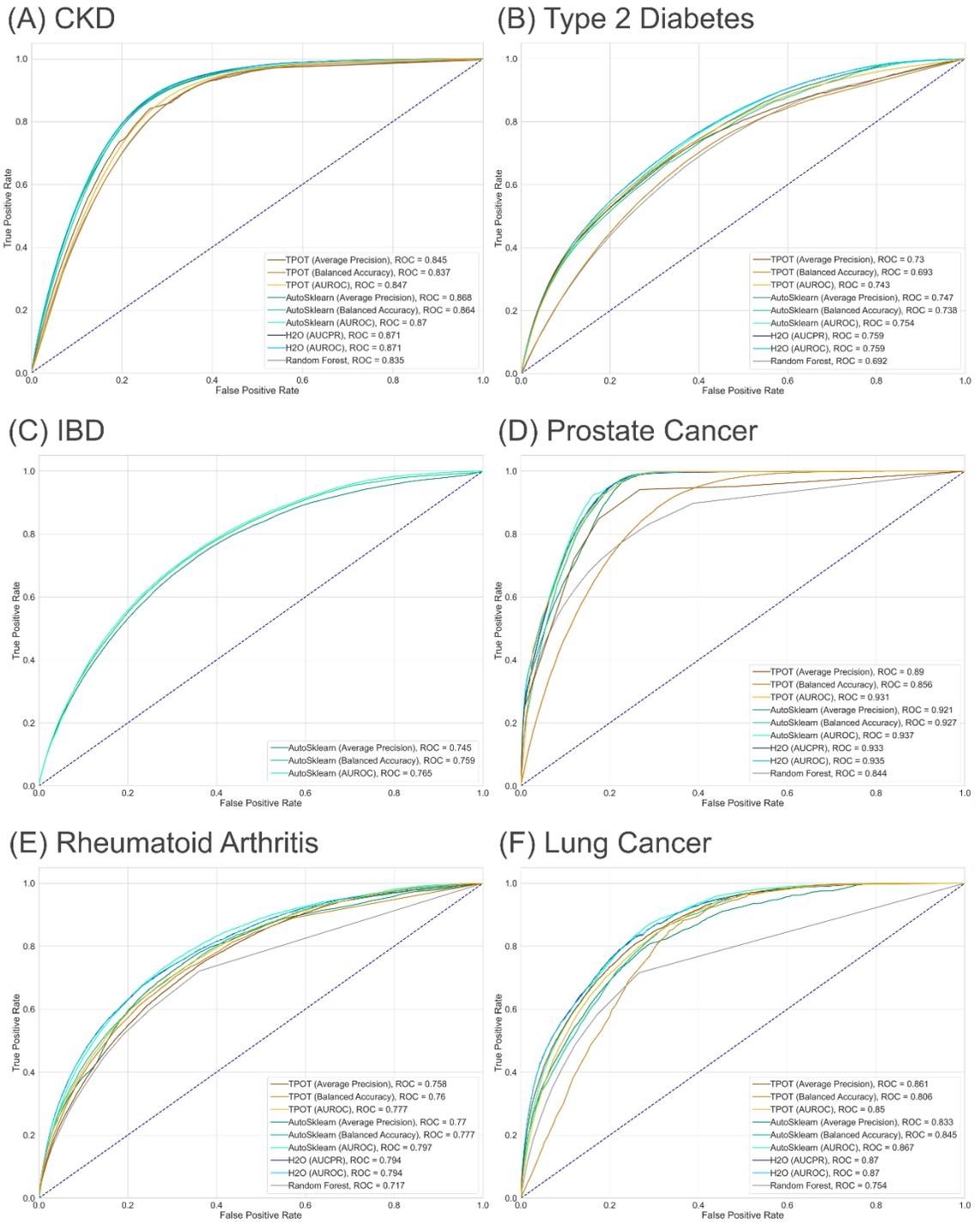

Figure 4: Receiver operating characteristic (ROC) curves of models trained for predicting different diseases. ROC curves are generated using prediction scores on full validation set (N = 12,125,832).

To illustrate, consider the case of predicting lung cancer, which has the lowest prevalence among the six diseases explored in this study. Lung cancer is often detected at the advanced stage when prognosis is poor and survival rates are low, thus making it one of the leading causes of cancer-related deaths in the United States. Several strategies that aim to detect the disease at an early stage where intervention is most effective are in place, chief of which are the rule-based screening guidelines provided by the National Comprehensive Cancer Network (NCCN) and the United States Preventive Services Task Force (USPSTF). However, even with these methods in place, only about 2% of annual lung cancer incidences are detected through screening. Patients who are considered eligible for screening based on the NCCN and USPSTF guidelines undergo a low-dose computed tomography (LDCT) annually. Though LDCT can detect lung cancer at a treatable stage, it also poses several health risks especially to those who are otherwise clear of the disease. These include unnecessary treatment, complications and a theoretical risk of developing cancer from exposure to low-dose radiation. Thus, in building a predictive model for lung cancer, these associated costs must be considered together with the objective of identifying as many positive cases as possible. In other words, for this kind of problem, there is a need to minimize the number of false positives while trying to achieve a high true positive rate (TPR). After training an AutoML model using any tool, caution should be exercised when still deploying the models. Models typically provide predictive probabilities and selecting the correct threshold for the application is necessary. Identifying the correct thresholds depending on the trade-offs between TPR and FPR can be done by looking at the respective ROC AUC curves as seen in Appendix Figure A3.

We show different confusion matrices for the best performing model for predicting lung cancer in terms of ROC AUC in Appendix Table A3. Thresholds are chosen based on deciles of actual predicted probability values for the full validation dataset. Identifying the optimal threshold will depend on the costs of true positives, false positives and false negatives. We consider hypothetical dollar costs for the same model noting that costs in terms of medical risks and quality of life are not included. We assume the per person cost of getting the disease is $300,000 annually if not detected early (equivalent to the cost of a false negative), while if detected early, the cost will be $84,000 annually (equivalent to the cost of a true positive). For this situation, we also consider two hypothetical tests, one priced at $100 per test and LDCT which costs about $500 on average. We compute savings based on the baseline situation where no tests are administered (each person with lung cancer is associated with the cost of a false negative). Figure 5 plots the savings for each hypothetical test cost per person for different decile probability thresholds. The optimal thresholds for the model depend on the situation where the model will be used. For the $100 test, we see the optimal cut-off is at the 70$^{th}$ percentile while for the $500 test, it is at the 90$^{th}$ percentile. For the $500 test, this cut-off is the only one that leads to positive savings. These cut-offs correspond to a FPR = 0.3, TPR = 0.9, and FPR = 0.1 and TPR = 0.52, respectively.

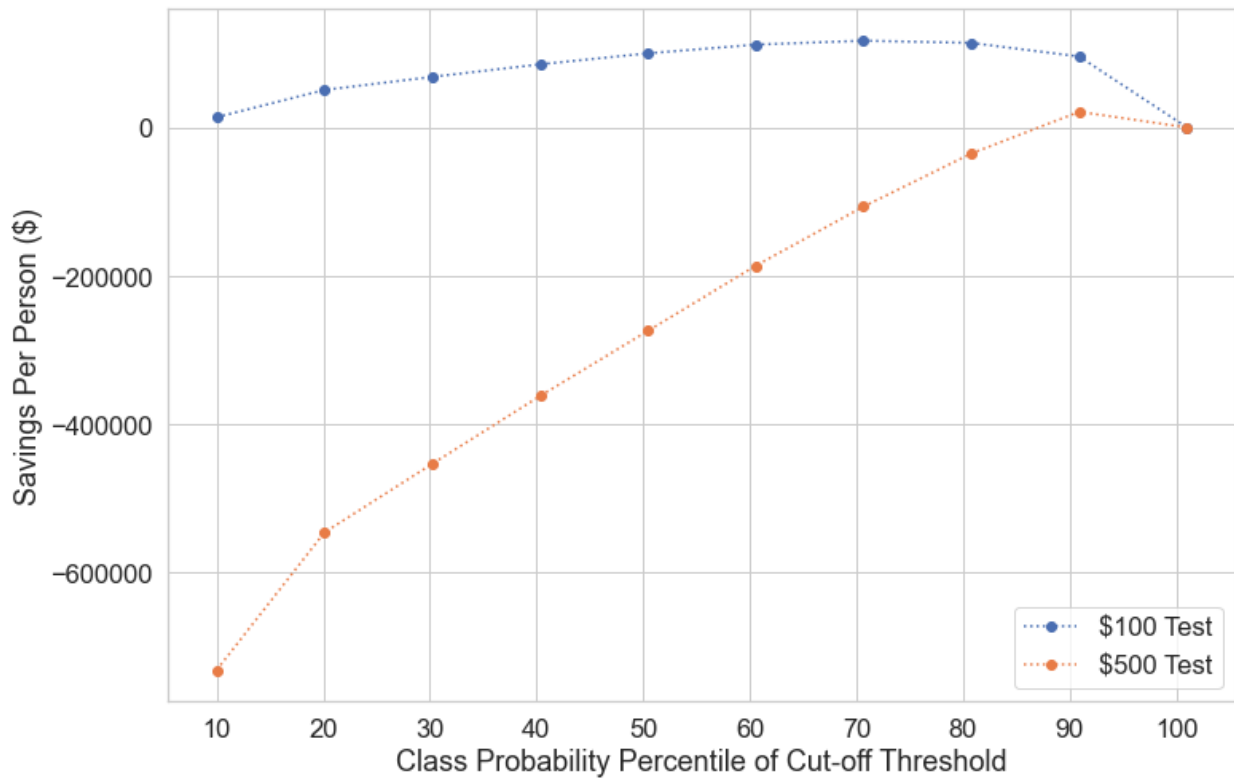

Figure 5: Average savings per person for different cut-off thresholds for the H2O (AUROC) model for different test costs. True positive costs are set at $84,000, while false negative costs are set at $300,000. False positive costs are only from the test costs.

## DISCUSSION

Since AutoML software packages are attractive out-of-the-box tools to build predictive models in the context of healthcare data, we examined and compared the performance of three of these tools (AutoSklearn, H2O, and TPOT) on a large medical claims dataset for six different disease outcomes. However, these datasets present several challenges. First, the sample size (~12.5M) is much larger than the typical size of datasets analyzed with AutoML. In this work we used a stratified sample of 300k for training, which is still quite large for AutoML given that these approaches are computationally intensive because they are iterating over many different algorithms. For example, the number of generations completed by TPOT within the 48-hour time limit varied greatly for each target disease and scoring metric. The number of generations completed ranged from from 7 to 38, with an average of 18.88 across 18 models. Improvements in terms of scalability of these AutoML methods are certainly desirable in the context of medical claims data. Once training several AutoML models each on a different and relatively large subsample of the dataset becomes computationally feasible, combining the resulting models into an ensemble may provide further performance improvements.

A second challenge is the extremely low case prevalence characteristic of healthcare data; in our examples, this varied from 0.053% to 0.63%. This may be the main culprit for the low AUCPR scores we observed across the methods and diseases. Improvements in terms of handling highly imbalanced datasets are crucial for healthcare applications. One direction for future work is to explore combinations of over- and under-sampling techniques with ensemble approaches in the spirit of [29].

Another challenge which may partly account for the poor performances observed among the models stems from the limitations inherent to the features available in healthcare databases. Since claims are coded for billing purposes, some healthcare services are tied to a certain ICD-10 code which may not necessarily be indicative of the presence of a certain disease. For example, individuals who are eligible for cancer screening will have the screening procedure billed under a cancer ICD-10 code regardless of the result. Hence, individuals who do not have cancer will still have cancer codes in their claims history. This means that simply flagging the presence of these ICD-10 codes is not an accurate representation of the person's medical history. Using fewer selected features may help improve model performance. For example, retaining only features corresponding to ICD-10 codes clinically related to the disease being predicted can reduce the size of the feature set and allow the models to more easily establish relationships between the features and the target.

## CONCLUSION

AutoML tools generally fast track the ML pipeline and the models they generate can serve as starting points for building predictors. However, the performance of these tools on the medical claims datasets used in this study suggest that there may be room for improvement in how AutoML tools handle data of this scale and with such high imbalance. To address the limitations of the data, further feature selection, resampling and imbalance-learning ensembles are possible next steps.

Despite the advantages of using AutoML tools for model selection and optimization, care must still be taken in identifying the optimal output thresholds depending on the research question.

## FUNDING

JM and EM were supported by the National Institutes of Health under award number LM010098.

## CONFLICTS OF INTEREST

Roland Albert Romero, Mariefel Nicole Deypalan, Suchit Mehrotra, John Titus Jungao, and Natalie Sheils are employees of OptumLabs part of UnitedHealth Group. Natalie Sheils owns stock in the company. The other authors have no conflicts of interest to disclose.

# APPENDIX

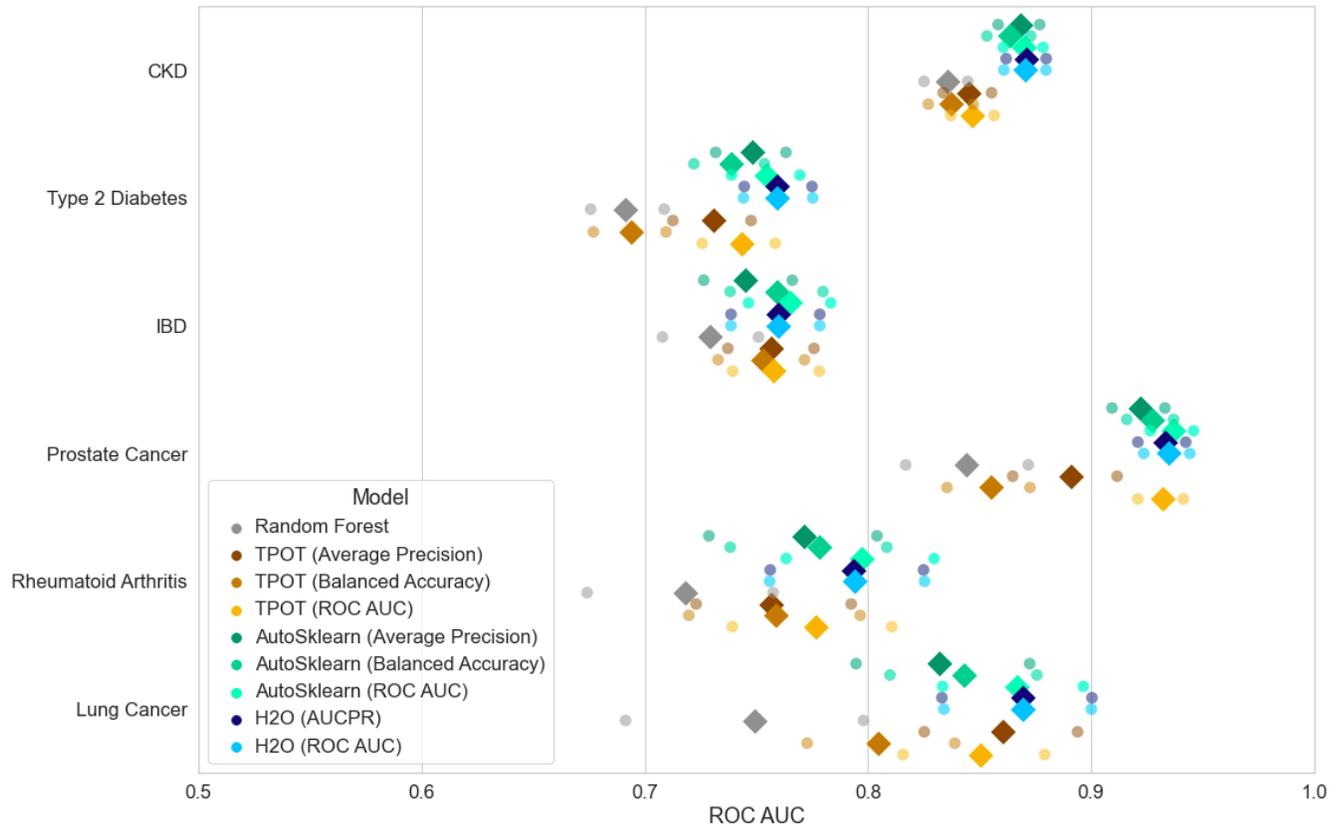

Figure A1: ROC AUC performance of different AutoML models trained for various disease outcomes from non-stratified bootstrap samples. Median values are indicated by diamond markers and 95% CI limits are indicated by circles.

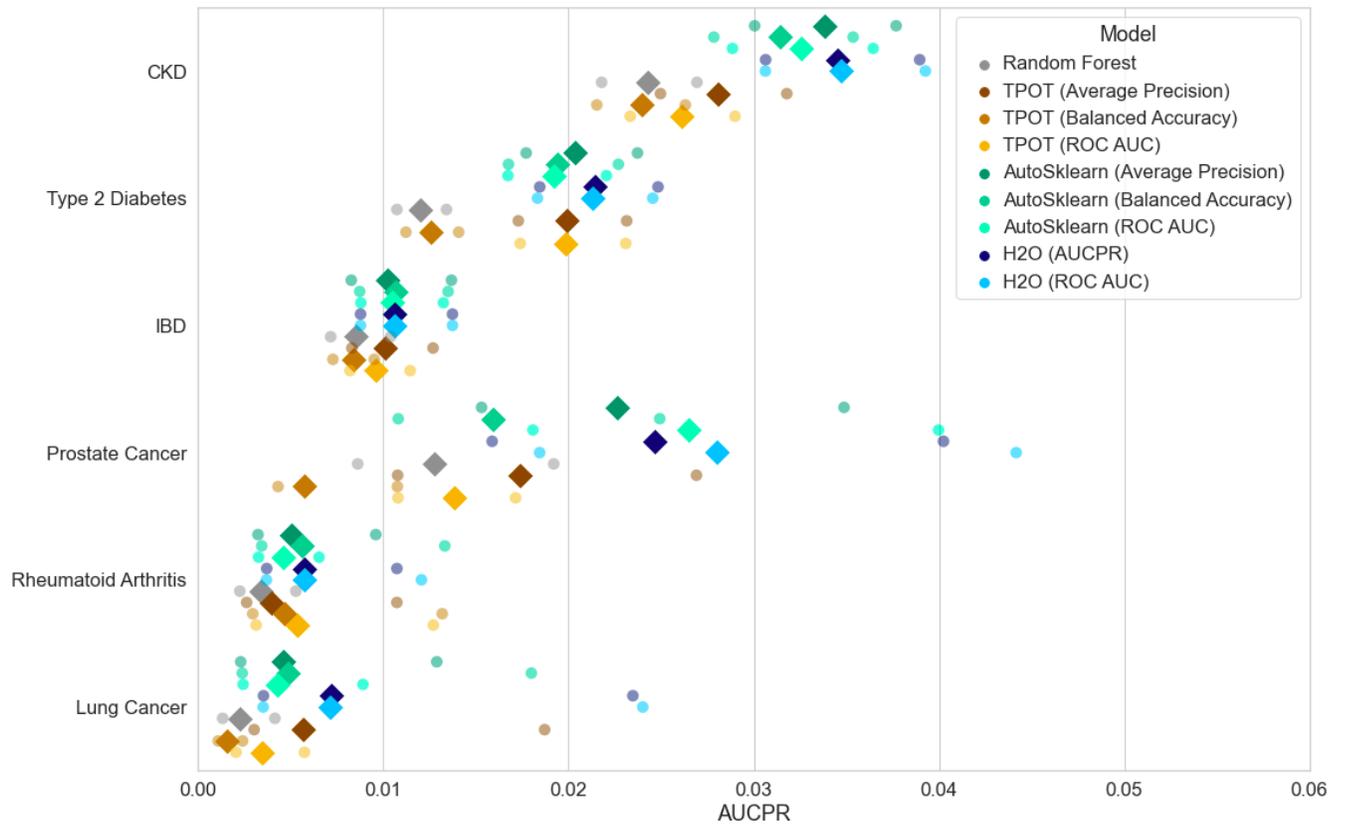

Figure A2: AUCPR performance of different AutoML models trained for various disease outcomes from non-stratified bootstrap samples. Median values are indicated by diamond markers and 95% CI limits are indicated by circles.

Table A1. Median ROC AUC performance (95% CI) of different AutoML models determined from stratified bootstrap samples of predictions on the validation set. Models with the best performance for each disease are indicated in bold.

|  | AutoSklearn (Average Precision) | AutoSklearn (Balanced Accuracy) | AutoSklearn (ROC AUC) | H2O (AUC) | H2O (AUCPR) | TPOT (Average Precision) | TPOT (Balanced Accuracy) | TPOT (ROC AUC) | Random Forest |
|---|---|---|---|---|---|---|---|---|---|
| CKD | 0.8685 (0.8604, 0.876) | 0.8638 (0.855, 0.8721) | 0.8702 (0.8623, 0.8782) | 0.8712 (0.8632, 0.879) | **0.8714 (0.8631, 0.8794)** | 0.8455 (0.8347, 0.8555) | 0.8374 (0.8279, 0.8475) | 0.847 (0.8382, 0.8562) | 0.8358 (0.8257, 0.8455) |
| Type 2 Diabetes | 0.7466 (0.7307, 0.7623) | 0.7382 (0.7212, 0.7526) | 0.7538 (0.7385, 0.769) | **0.7585 (0.7425, 0.7725)** | 0.7584 (0.743, 0.7728) | 0.7295 (0.7112, 0.7462) | 0.6926 (0.6752, 0.7087) | 0.7425 (0.7252, 0.7582) | 0.6907 (0.6753, 0.7064) |
| IBD | 0.745 (0.7239, 0.7646) | 0.7597 (0.7385, 0.7775) | 0.7642 (0.7446, 0.7825) | **0.7593 (0.7399, 0.7773)** | **0.7593 (0.7399, 0.7773)** | 0.7563 (0.7381, 0.7746) | 0.7522 (0.7341, 0.7702) | 0.7578 (0.7373, 0.7763) | 0.729 (0.7068, 0.7495) |
| Prostate Cancer | **0.9213 (0.9088, 0.9323)** | 0.9266 (0.9166, 0.9366) | **0.9369 (0.9254, 0.9463)** | 0.9347 (0.9233, 0.9448) | 0.9328 (0.9216, 0.9425) | 0.8897 (0.8627, 0.9094) | 0.8553 (0.8378, 0.8739) | 0.9315 (0.9202, 0.9415) | 0.8447 (0.814, 0.8721) |
| Rheumatoid Arthritis | 0.7699 (0.7346, 0.8087) | 0.7771 (0.7433, 0.8143) | 0.7974 (0.7629, 0.8288) | **0.7952 (0.7588, 0.8275)** | 0.7948 (0.7576, 0.8275) | 0.7578 (0.7213, 0.7931) | 0.7599 (0.7236, 0.7961) | 0.7788 (0.7416, 0.8118) | 0.7184 (0.6714, 0.7579) |
| Lung Cancer | 0.8325 (0.7877, 0.8725) | 0.8448 (0.8075, 0.8757) | 0.8665 (0.8333, 0.8996) | **0.8712 (0.8366, 0.9067)** | 0.8711 (0.8365, 0.9071) | 0.8602 (0.8261, 0.8933) | 0.8055 (0.7726, 0.8394) | 0.8485 (0.8155, 0.886) | 0.7519 (0.7003, 0.8124) |

Table A2. Median AUCPR performance (95% CI) of different AutoML models determined from stratified bootstrap samples of predictions on the validation set. Models with the best performance for each disease are indicated in bold.

|  | AutoSklearn (Average Precision) | AutoSklearn (Balanced Accuracy) | AutoSklearn (ROC AUC) | H2O (AUC) | H2O (AUCPR) | TPOT (Average Precision) | TPOT (Balanced Accuracy) | TPOT (ROC AUC) | Random Forest |
|---|---|---|---|---|---|---|---|---|---|
| **CKD** | 0.0339 (0.0309, 0.038) | 0.0316 (0.0289, 0.0352) | 0.0325 (0.0298, 0.0359) | **0.0347 (0.0317, 0.0384)** | 0.0345 (0.0313, 0.0386) | 0.0282 (0.0257, 0.0314) | 0.0239 (0.0223, 0.0256) | 0.0261 (0.0239, 0.029) | 0.0243 (0.0223, 0.0264) |
| **Type 2 Diabetes** | 0.0201 (0.0174, 0.0231) | 0.0193 (0.0167, 0.0221) | 0.0191 (0.0171, 0.0216) | 0.021 (0.0185, 0.0244) | **0.0212 (0.0186, 0.0246)** | 0.0198 (0.0172, 0.0226) | 0.0124 (0.0113, 0.0138) | 0.0198 (0.0173, 0.0224) | 0.012 (0.0109, 0.013) |
| **IBD** | 0.0103 (0.0086, 0.0131) | 0.0107 (0.009, 0.0138) | 0.0105 (0.009, 0.013) | **0.0107 (0.009, 0.0135)** | **0.0107 (0.009, 0.0135)** | 0.0102 (0.0086, 0.0124) | 0.0083 (0.0075, 0.0092) | 0.0096 (0.0085, 0.0111) | 0.0085 (0.0074, 0.0099) |
| **Prostate Cancer** | 0.0227 (0.0155, 0.0324) | 0.016 (0.0114, 0.0242) | 0.0267 (0.0181, 0.0396) | **0.0281 (0.0189, 0.0415)** | 0.0249 (0.0168, 0.038) | 0.0171 (0.0114, 0.025) | 0.0058 (0.0047, 0.0107) | 0.0138 (0.0113, 0.0165) | 0.0127 (0.009, 0.0182) |
| **Rheumatoid Arthritis** | 0.005 (0.0033, 0.0094) | 0.0056 (0.0036, 0.0129) | 0.0046 (0.0035, 0.0062) | **0.0058 (0.004, 0.0105)** | **0.0058 (0.004, 0.0103)** | 0.0039 (0.0027, 0.0087) | 0.0045 (0.003, 0.0116) | 0.0053 (0.0034, 0.0136) | 0.0034 (0.0023, 0.0052) |
| **Lung Cancer** | 0.0045 (0.0024, 0.0122) | 0.0047 (0.0025, 0.0149) | 0.0043 (0.0026, 0.0084) | 0.0074 (0.0038, 0.0216) | **0.0075 (0.0038, 0.0211)** | 0.0059 (0.0032, 0.0189) | 0.0016 (0.0012, 0.0023) | 0.0035 (0.0023, 0.0058) | 0.0023 (0.0014, 0.0041) |

Table A3. Confusion matrices for H2O (AUROC) model for predicting lung cancer for thresholds set at decile values of validation set prediction probabilities.

| Percentile | Threshold | TN | FP | FN | TP | FPR | TPR |
| --- | --- | --- | --- | --- | --- | --- | --- |
| 10 | 0.000050 | 1332 | 12118119 | 0 | 6381 | 1.000 | 1.000 |
| 20 | 0.000063 | 2420542 | 9698909 | 13 | 6368 | 0.800 | 0.998 |
| 30 | 0.000076 | 3637704 | 8481747 | 46 | 6335 | 0.700 | 0.993 |
| 40 | 0.000100 | 4850231 | 7269220 | 101 | 6280 | 0.600 | 0.984 |
| 50 | 0.000148 | 6038573 | 6080878 | 211 | 6170 | 0.502 | 0.967 |
| 60 | 0.000247 | 7275056 | 4844395 | 443 | 5938 | 0.400 | 0.931 |
| 70 | 0.000412 | 8486907 | 3632544 | 868 | 5513 | 0.300 | 0.864 |
| 80 | 0.000627 | 9699105 | 2420346 | 1559 | 4822 | 0.200 | 0.756 |
| 90 | 0.001109 | 10910511 | 1208940 | 2737 | 3644 | 0.100 | 0.571 |
| 100 | 0.061339 | 12119450 | 1 | 6381 | 0 | 0.000 | 0.000 |

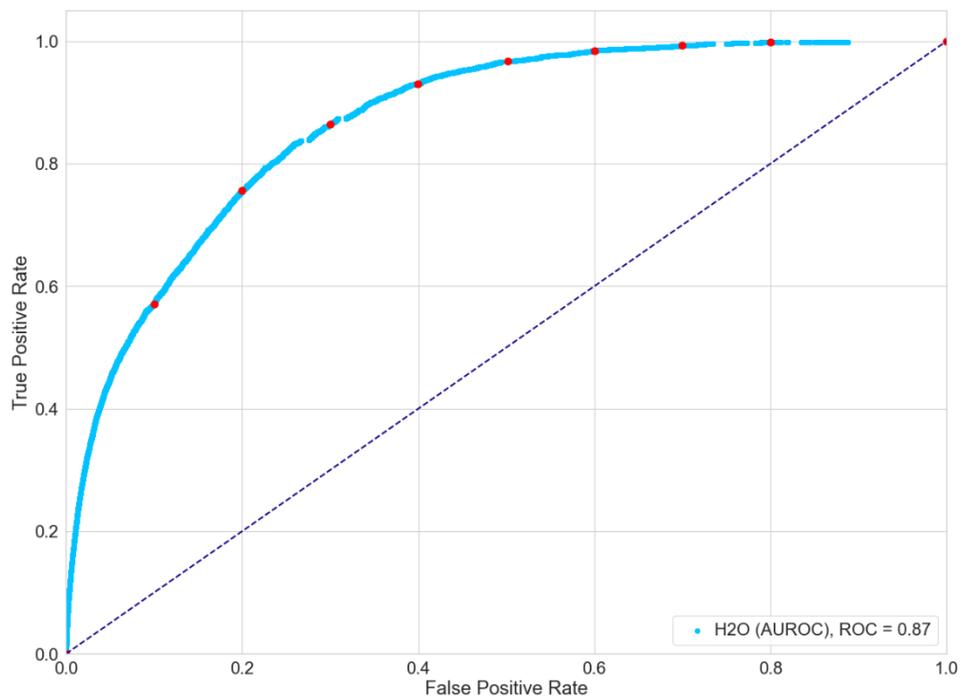

Figure A3: ROC curve of H2O (AUROC) model for predicting lung cancer on validation set. Red dots indicate locations of decile probabilities.